\documentclass[sigconf]{acmart}
\usepackage{xcolor}
\AtBeginDocument{%
  }

\copyrightyear{2026}
\acmYear{2026}
\setcopyright{cc}
\setcctype{by}
\acmConference[CIKM '26]{Proceedings of the 35th ACM International Conference on Information and Knowledge Management}{November 07--11, 2026}{Rome, Italy}
\acmBooktitle{Proceedings of the 35th ACM International Conference on Information and Knowledge Management (CIKM '26), November 07--11, 2026, Rome, Italy}
\acmDOI{10.1145/3799682.3841039}
\acmISBN{979-8-4007-2539-5/2026/11}
\newif\iftamreviewmarkup
\tamreviewmarkupfalse
\newcommand{\reviewtag}[1]{}
\newcommand{\rev}[2]{\iftamreviewmarkup\textcolor{blue}{#2}\reviewtag{#1}\else #2\fi}
\definecolor{tamagentcolor}{RGB}{0,110,90}
\newcommand{\agentrev}[1]{\iftamreviewmarkup\textcolor{tamagentcolor}{#1}\else #1\fi}
\newenvironment{tampromptexcerpt}{%
  \begin{list}{}{%
    \setlength{\leftmargin}{0.6em}%
    \setlength{\rightmargin}{0pt}%
    \setlength{\topsep}{2pt}%
    \setlength{\partopsep}{0pt}%
    \setlength{\parsep}{0pt}%
    \setlength{\itemsep}{0pt}%
  }%
  \item\small
}{%
  \end{list}%
}

\begin{document}

\title{Tasks over Application Manuals: Revealing Gaps in Long-Horizon Procedural Reasoning for Language Models}

\author{Utkarsh Soni}
\correspondingauthor
\orcid{0000-0002-1588-8861}
\email{utkarsh\_soni@manulife.com}
\affiliation{%
  \institution{Manulife}
  \city{Pittsburgh}
  \country{United States}
}

\author{Syed Shariyar Murtaza}
\orcid{0000-0003-3330-4783}
\email{syed\_shariyar\_murtaza@manulife.com}
\affiliation{%
  \institution{Manulife}
  \city{Toronto}
  \country{Canada}
}

\author{Yifan Nie}
\orcid{0009-0006-9254-4735}
\email{yifan\_nie@manulife.com}
\affiliation{%
  \institution{Manulife}
  \city{Toronto}
  \country{Canada}
}

\author{Sachin Chandrasekhar}
\orcid{0009-0009-6081-1431}
\email{sachin\_chandrasekhar@manulife.com}
\affiliation{%
  \institution{Manulife}
  \city{Bangalore}
  \country{India}
}

\author{Eugene Wen}
\orcid{0009-0002-7683-3693}
\email{eugene\_wen@manulife.com}
\affiliation{%
  \institution{Manulife}
  \city{Toronto}
  \country{Canada}
}

\renewcommand{\shortauthors}{Utkarsh Soni, Syed Shariyar Murtaza, Yifan Nie, Sachin Chandrasekhar, and Eugene Wen}

\begin{abstract}
  Large language models (LLMs) have achieved strong performance on a wide range of natural language tasks, and recent benchmarks suggest that they are increasingly adept at multi-hop reasoning. However, these benchmarks are typically short-horizon, requiring only a small number of retrieval or inference steps, and provide limited evidence of reliability on real-world tasks that involve following manuals spanning hundreds of pages with complex, interdependent guidelines. In this paper, we introduce \emph{Tasks over Application Manuals} (TAM), a benchmark for evaluating long-horizon procedural reasoning. We construct TAM by curating real-world tasks from two domains—ICD-10-CM clinical coding (mapping medical conditions to diagnostic codes) and U.S. federal sentencing (computing crime sentencing guideline outcomes, specifically offense levels)—with human-validated labels. Each task requires following an authoritative manual with tens of thousands of rules and executing a sequence of interdependent steps across different sections to produce an exact answer. \agentrev{We evaluate general-purpose prompting approaches, including retrieval-augmented generation, ReAct-style prompting, and an agent-harness baseline on GPT-5, and find that the best exact-match performance remains extremely low: 1\% on ICD-10-CM coding and 15.5\% on sentencing tasks.} These results show that current benchmarks may overestimate LLM reasoning ability and miss a key challenge: reliably following long, rule-based procedures. The complete TAM data and code are publicly available.\footnote{Data: \url{https://huggingface.co/datasets/manulife/tam-benchmarks}. Code: \url{https://github.com/manulife-ai/tasks-over-application-manual}.}

\end{abstract}



\maketitle

\section{Introduction}

Large language models (LLMs) have significantly advanced natural language processing, achieving strong performance across tasks such as question answering, summarization, and code generation. Alongside these advances, there has been growing interest in evaluating whether LLMs can perform reasoning, with several benchmarks suggesting strong performance on multi-step and multi-hop problems.

However, these benchmarks are limited in scope. Many widely used reasoning tasks, such as HotpotQA~\citep{yang2018hotpotqa} and MuSiQue~\citep{trivedi2021musique}, are short-horizon in nature, typically requiring only a small number of retrieval or inference steps. While useful, such tasks do not capture an important aspect of real-world reasoning: following long, structured procedures defined by external manuals.

In many real-world and industrial settings, solving a task means following a step-by-step process described over tens of thousands of guidelines in a large document. For example, in ICD-10-CM coding, the coder starts with a clinical description, looks up candidate medical conditions, checks rules across dozens of sections, and often loops back to refine earlier choices based on identified constraints. In federal sentencing, a user starts from the crime, identifies the relevant guideline, applies a sequence of adjustments, and aggregates decisions across multiple charges to compute a final sentencing range. These are not one-shot decisions—they often involve dozens of intermediate steps where each choice affects what comes next, making the overall process highly sensitive to earlier mistakes.

This difference changes the nature of the problem. Instead of combining a few pieces of evidence, the system must maintain consistency across a sequence of decisions, follow ordering rules, and apply constraints correctly throughout the process. Small mistakes early on can invalidate later steps, even if each intermediate decision appears reasonable in isolation. These challenges are not directly tested in existing benchmarks, which focus on shorter reasoning chains. This gap motivates the need for benchmarks that test whether models can reliably follow long, rule-based procedures.

To address this, we introduce \emph{Tasks over Application Manuals} (TAM), a benchmark for long-horizon procedural reasoning. TAM is built from real-world tasks that require following an authoritative manual. Unlike existing benchmarks that require a small number of hops, TAM involves dozens of interdependent decisions across manuals with tens of thousands of rules and cross-references, where correctness depends on executing the entire procedure without error.

We curated two such application manuals: ICD-10-CM clinical coding and federal sentencing guideline calculation. The ICD benchmark is grounded in the 2019 ICD-10-CM Guidelines, the 1{,}304-page Alphabetic Index, and the 1{,}942-page Tabular List. Our dataset consists of 1{,}000 cases sampled from 38{,}332 eligible encounters of real ICD coding.

The legal benchmark, on the other hand, consists of five annual Title~18 archives and five annual USSG manuals spanning 2021--2025, where the data comprise 200 approved cases after human review. Across both domains, the primary difficulty lies in following the manual end-to-end, typically involving the retrieval of several relevant passages.

On this benchmark, we evaluate single-pass RAG, agentic RAG, ReAct-style tool use, \agentrev{and an agent-harness baseline} on GPT-5, all of which are representative techniques that perform well on existing multi-hop reasoning benchmarks. \rev{MR-3,R1-5,R2-1,R3-3}{We treat these experiments as strong initial reference baselines for a new benchmark rather than as an exhaustive comparison against every specialized or newly released method.} \agentrev{We find that they struggle on both TAM tasks, with exact match remaining at 1\% or below for ICD-10-CM coding and reaching at most 15.5\% for sentencing.} This highlights a clear gap between short-horizon reasoning performance and real-world procedural execution.

Our contributions are as follows:
\begin{itemize}
\item We introduce \textbf{Tasks over Application Manuals (TAM)}, a benchmark for long-horizon procedural reasoning built from real-world domains—ICD-10-CM clinical coding and U.S. federal sentencing—where solving tasks requires following complex, multi-step procedures defined across large, structured manuals.

\item We formalize \textbf{manual-based reasoning} as a distinct evaluation setting, where correctness depends on maintaining consistency across dozens of interdependent decisions, rather than retrieving or combining a small set of relevant facts.

\item We evaluate representative GPT-5 baselines—single-pass RAG, agentic RAG, ReAct-style tool use, \agentrev{and an agent-harness baseline}—and show that \agentrev{they struggle on TAM, with exact match remaining at 1\% or below on ICD-10-CM coding and reaching at most 15.5\% on sentencing.}

\item We identify a critical gap between performance on existing short-horizon reasoning benchmarks and real-world procedural tasks, highlighting the need for new methods that can reliably execute long, rule-based processes end-to-end.
\end{itemize}

\section{Tasks over Application Manuals}

\subsection{Task Definition}

We define a \emph{Task over Application Manuals} (TAM) instance as a triple $\left(x, \mathcal{M}, y^{*}\right)$, where $x$ is a case description, $\mathcal{M}$ is a manual, and $y^{*}$ is the exact output obtained by correctly applying the guidelines present in the manual to the case. The central object is not a single retrieved passage but a \emph{decision-trajectory}. Starting from an initial state $s_0 = (x, \emptyset, \emptyset)$, a system iteratively selects actions $a_t \in \mathcal{A}$ and updates its state $s_{t+1} = \tau_{\mathcal{M}}(s_t, a_t)$, where $s_t = (x, C_t, y_t)$ represents the accumulated context and partial output. Here, $C_t$ denotes the global state, including retrieved content, intermediate commitments, unresolved subproblems, and active constraints, while $y_t$ is the partial output constructed so far at step $t$.

The transition operator $\tau_{\mathcal{M}}$ specifies how actions interact with the manual. The action space $\mathcal{A}$ includes retrieving rules, resolving cross-references, applying constraints, refining outputs, planning over subgoals, revising earlier decisions, backtracking, and termination. A TAM instance is solved only if the system halts at some step $H$ with $y_H = y^{*}$ and the resulting trajectory remains admissible under the manual throughout execution.

The key challenge is that local plausibility does not imply global validity. At any step, several actions may look reasonable given the current evidence, yet only a small subset preserve a path to a valid final output. A missed exception, premature commitment, or unsupported assumption can silently eliminate the correct solution and only become visible several steps later. Solving TAM is therefore a constrained search problem over partial solutions: the system must plan, track active obligations, detect when the current path is inconsistent, and revise or backtrack before termination. Retrieval helps expose relevant rules, but it does not by itself ensure that the evolving output is complete, correctly ordered, and globally valid.

\begin{figure*}[t]
\centering
\begin{minipage}[t]{0.40\textwidth}
\centering
\includegraphics[width=\linewidth]{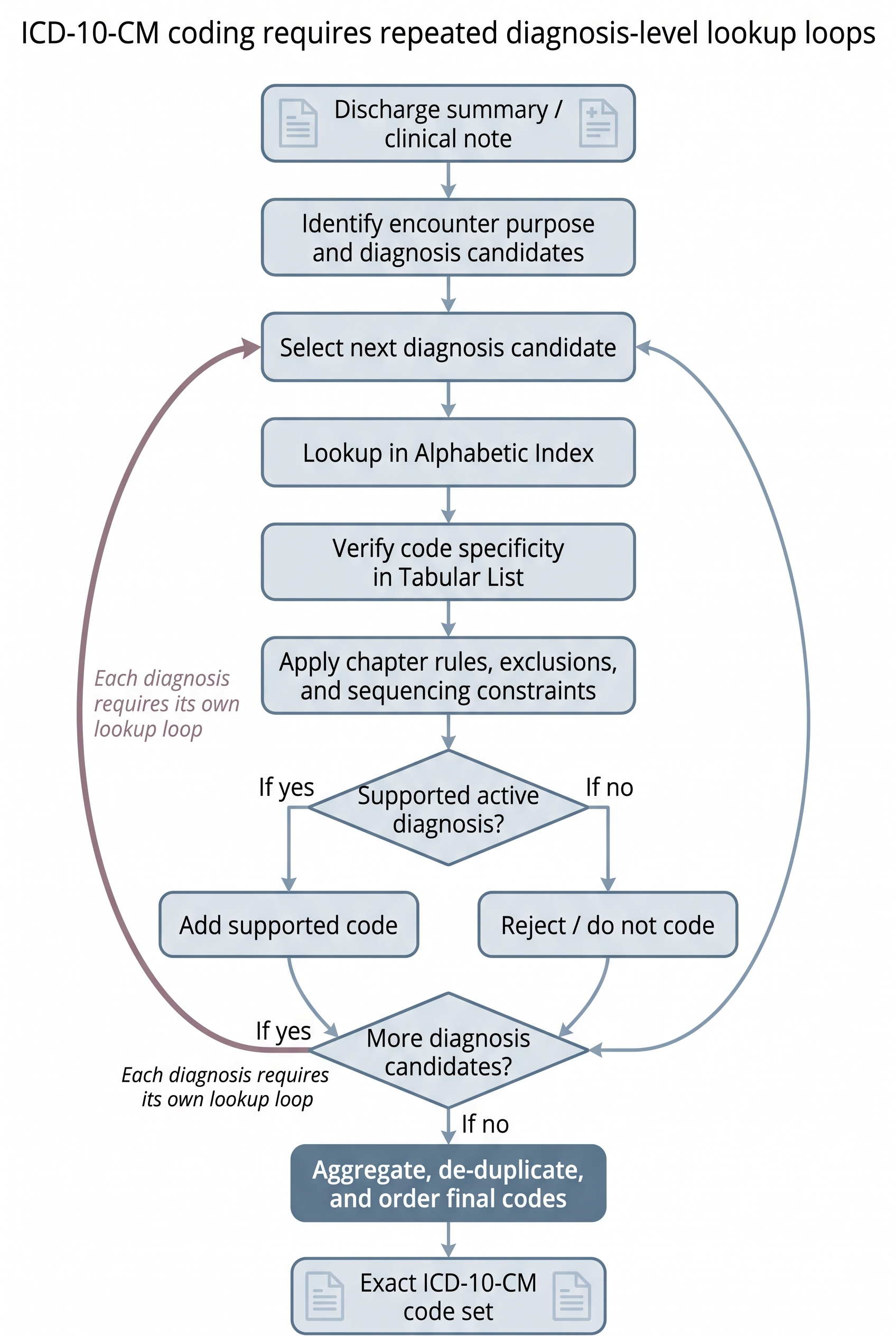}
\smallskip

\textbf{(a) ICD-10-CM clinical coding.}
\end{minipage}\hfill
\begin{minipage}[t]{0.40\textwidth}
\centering
\includegraphics[width=\linewidth]{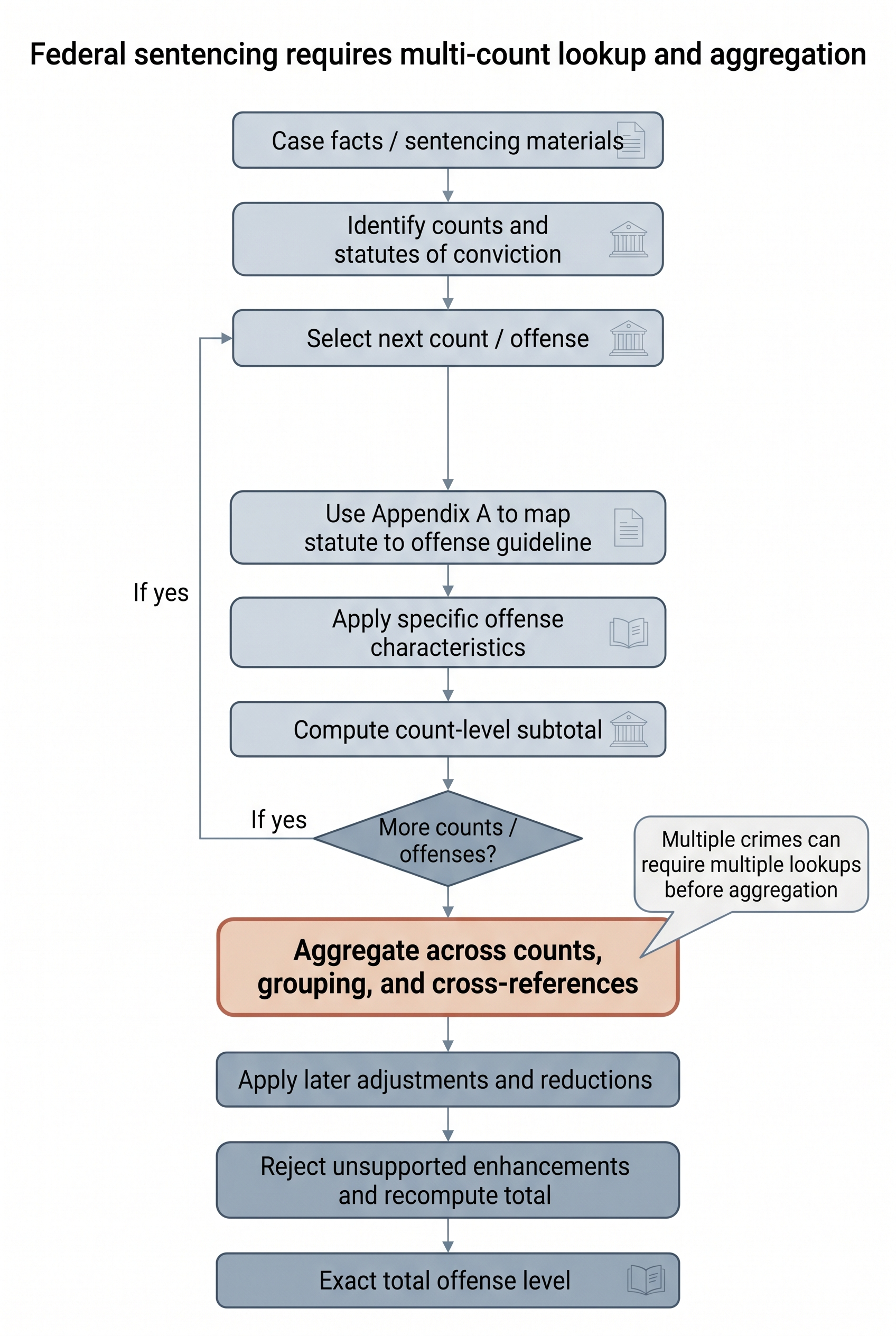}
\smallskip

\textbf{(b) Federal sentencing guideline calculation.}
\end{minipage}
\caption{Conceptual execution patterns in the two TAM domains. In ICD-10-CM coding, each diagnosis candidate can trigger its own lookup loop through the Alphabetic Index, the Tabular List, and guideline constraints before the final code set is aggregated, de-duplicated, and ordered. In federal sentencing, multiple counts or offenses can require repeated guideline lookups and count-level subtotal calculations before the system aggregates across counts, grouping logic, cross-references, and later adjustments to reach the exact total offense level.}
\Description{Two side-by-side flowcharts. The left flowchart shows ICD-10-CM coding as a repeated diagnosis-level lookup loop: a discharge summary leads to encounter and diagnosis identification, then the coder repeatedly selects the next diagnosis candidate, looks it up in the Alphabetic Index, verifies specificity in the Tabular List, applies chapter rules and sequencing constraints, decides whether the diagnosis is supported, and either adds or rejects the code before looping to the next diagnosis candidate. When no candidates remain, the coder aggregates, de-duplicates, and orders the final ICD code set. The right flowchart shows federal sentencing as a repeated count-level lookup and aggregation process: case facts lead to identifying counts and statutes, then the user selects the next count or offense, uses Appendix A to map the statute to the offense guideline, applies specific offense characteristics, computes a count-level subtotal, loops over additional counts or offenses when needed, then aggregates across counts, grouping rules, and cross-references before applying later adjustments and reductions, rejecting unsupported enhancements, and producing the exact total offense level.}
\label{fig:tam-domain-flows}
\end{figure*}

Figure~\ref{fig:tam-domain-flows} makes this procedural burden concrete in our two domains. In ICD-10-CM coding, each diagnosis candidate can trigger its own lookup loop. In federal sentencing, the system may need to repeat count-level lookups and then aggregate the resulting subtotals under grouping, cross-reference, and later-adjustment rules.

\subsection{Illustrative Example: ICD-10-CM Clinical Coding}

ICD-10-CM \footnote[1]{https://www.cdc.gov/nchs/icd/icd-10-cm/index.html} clinical coding provides a concrete instantiation of the TAM setting. Here, the input $x$ is a hospital discharge summary, often a dense narrative that mixes the reason for admission, active diagnoses, complications, chronic conditions, medications, and copied background history. The manual $\mathcal{M}$ consists of multiple interacting resources: a term-to-code lookup, a structured code hierarchy, and guidelines that specify inclusion criteria, exclusions, sequencing rules, and when additional codes are required for any diagnosed code. The output $y^{*}$ is the exact reportable ICD-10-CM code set together with its required sequencing information. In evaluation, we score exact match of the code set ignoring order and report primary-diagnosis accuracy separately. Solving the task therefore means converting a messy narrative into a complete and rule-consistent coding decision --- something that goes beyond simply retrieving a few relevant passages for an answer.

Consider a patient admitted for chemotherapy for metastatic cancer. The same record may also mention liver metastases, a drug reaction during infusion, hypertension and depression managed during the stay, and additional conditions copied from prior history sections. The system cannot code these facts independently. It must determine what the encounter is primarily for, whether the treatment complication introduces another required code, which chronic conditions are active enough to include, which mentions are background context rather than billable diagnoses for this encounter, and how the final codes must be ordered. One note can contain many diagnosis-like statements, but the output must be exact rather than exhaustive. The source of all these rules is the manual which must be frequently consulted for the final decision. 

This creates characteristic failure modes. A system that codes only the cancer misses the encounter-for-chemotherapy code that the guidelines require to be sequenced first (this first-listed code is called the primary diagnosis). A system that codes every diagnosis-like phrase produces an overinclusive list. A system that retrieves the right rule but attaches it to the wrong primary problem can produce reasoning that looks locally coherent while still yielding a globally invalid answer. We chose this domain in TAM to expose such failure modes. 

\section{Related Work}

\subsection{Retrieval and short-horizon reasoning benchmarks}

Information retrieval and question answering benchmarks evaluate how well systems locate and combine relevant evidence. Datasets such as MS MARCO~\cite{nguyen2016msmarco}, BEIR~\cite{thakur2021beir}, and the TREC Deep Learning track~\cite{craswell2020overview} focus on document ranking, while multi-hop QA and fact verification benchmarks such as HotpotQA~\cite{yang2018hotpotqa}, MuSiQue~\cite{trivedi2021musique}, Natural Questions~\cite{kwiatkowski2019natural}, FEVER~\cite{thorne2018fever}, and KILT~\cite{petroni2021kilt} require aggregating evidence across multiple sources.

These benchmarks have driven major progress, but they remain fundamentally short-horizon: most instances can be solved by combining a small number of facts. Evaluation emphasizes retrieval relevance or final answer correctness rather than whether intermediate decisions form a valid procedure. As a result, they provide limited visibility into failures that emerge only over longer, stateful reasoning trajectories.

\subsection{Procedural and constraint-based reasoning benchmarks}

Recent work has begun to study reasoning in interactive planning domains and under procedural or constraint-heavy settings. Interactive benchmarks such as WebShop~\cite{yao2022webshop} and ALFWorld~\cite{shridhar2021alfworld} evaluate agents acting in an environment, where progress depends on exploration, observation, and environment feedback. ConstraintBench~\cite{constraintbench} and COMPASS~\cite{compass} evaluate multi-constraint reasoning in optimization and planning, while BioProBench~\cite{bioprobench} and ProcedureVQA~\cite{procedurevqa} assess step-level procedural reasoning. \agentrev{More recent benchmarks such as SWE-bench~\cite{jimenez2024swe}, AgentBench~\cite{liu2024agentbench}, and Agents' Last Exam~\cite{sun2026agents} extend this line of work to harder multi-step software and agent tasks, but they still do not center faithful execution against a fixed, cross-referenced professional manual.}

These benchmarks move beyond single-shot QA, but they usually emphasize environment interaction, abstract feasibility, or local transitions rather than executing a fixed procedure over an external, cross-referenced manual. TAM instead targets manual-grounded multi-step execution: the system must plan across dependent decisions, but the admissible actions and final outputs are specified by distributed source documents rather than discovered through open-ended exploration. In that sense, TAM sits between short-horizon multi-hop retrieval and interactive planning benchmarks, focusing on faithful execution of a complex written procedure.

\subsection{Domain-specific rule-governed tasks}

Many real-world tasks are governed by complex rule systems but are commonly framed as direct prediction problems. In clinical NLP, ICD coding is usually treated as extreme multi-label classification, where models map notes directly to codes~\cite{mullenbach-etal-2018-explainable,laat_vu2020,huang-etal-2022-plm}. In legal AI research, benchmarks such as CAIL2018~\cite{cail2018} similarly cast reasoning as predicting charges or outcomes from fact descriptions.

These formulations can be effective for output prediction, but they abstract away the procedural structure that practitioners actually follow. In practice, tasks such as ICD coding require navigating manuals, applying exclusions, checking specificity, and enforcing coding rules~\cite{johnson2016mimic,johnson2023mimiciv}; for LLMs, that manual-following capability is especially important where labeled data are scarce and direct prediction models are harder to build. Because the evaluation target is only the final output, these settings do not test whether a system can correctly execute the underlying rule-governed process.

\subsection{Retrieval-augmented and agentic approaches}

A large body of work extends language models with retrieval and tool use for knowledge-intensive tasks. Retrieval-augmented generation methods such as REALM~\cite{guu2020realm}, RAG~\cite{lewis2020rag}, FiD~\cite{izacard2021fid}, RETRO~\cite{borgeaud2022retro}, and Atlas~\cite{izacard2023atlas} improve access to external knowledge, while retrieval models such as DPR~\cite{karpukhin2020dense} and ColBERT~\cite{khattab2020colbert} improve evidence quality. Agentic systems such as ReAct~\cite{yao2023react}, Toolformer~\cite{schick2023toolformer}, and WebGPT~\cite{nakano2021webgpt} further support iterative reasoning and tool interaction. \agentrev{More recent agent harnesses, such as Claude Code~\cite{anthropic_claude_code2026}, the OpenAI Agents SDK~\cite{openai_agents_sdk2026}, and LangChain Deep Agents~\cite{langchain_deepagents2026}, add an orchestration layer over the same model-and-tool loop, supporting planning, reusable instructions or skills, delegation to specialized agents that often work in their own local context windows rather than one shared growing conversation, and the passing of intermediate results across long tasks.}

In this paper, we focus on representative paradigms---standard RAG, agentic retrieval, and ReAct-style reasoning, \agentrev{and an agent-harness baseline}---as strong prompting-based reference baselines. These methods perform well on existing retrieval and short-horizon reasoning benchmarks, but they are not designed to preserve consistency over long, manual-governed decision trajectories.

\subsection{Positioning of TAM}

TAM targets a distinct capability: manual-governed procedural execution. Unlike retrieval and QA benchmarks, it evaluates whether a system maintains a valid decision trajectory rather than merely identifying relevant evidence. Unlike procedural and constraint benchmarks, it requires reasoning over structured, cross-referenced manuals. And unlike domain-specific prediction tasks, it targets a capability that can be exercised from manuals and case inputs directly, rather than one that depends on large task-specific training sets. In summary, the gap TAM isolates is straightforward: existing benchmarks test retrieval, short-horizon reasoning, or constraint satisfaction in isolation, but they do not assess whether models can execute extended, rule-governed procedures grounded in real manuals. TAM makes that capability explicit.

\section{Benchmark Construction}

We instantiate TAM in two professional workflows where a case record must be mapped to an exact output under a distributed manual: ICD-10-CM clinical coding and federal sentencing guideline calculation. In both domains, the manual governs the reasoning process: systems must follow a valid path through multiple interacting components and terminate with an exact answer. This section describes their structure, sources of difficulty, and some notes on how it was constructed. TAM data are available on Hugging Face\footnote{\url{https://huggingface.co/datasets/manulife/tam-benchmarks}.}, and code is available on GitHub.\footnote{\url{https://github.com/manulife-ai/tasks-over-application-manual}.}

\begin{table*}
\caption{Short task excerpts and brief manual cues from the two TAM tasks used in this paper. Full worked out examples appear in Appendix~\ref{app:icd-case-study} and Appendix~\ref{app:legal-case-study}.}
\label{tab:tam-excerpts}
\centering
\small
\setlength{\tabcolsep}{5pt}
\renewcommand{\arraystretch}{1.08}
\begin{tabular}{@{}p{0.16\textwidth}p{0.27\textwidth}p{0.18\textwidth}p{0.31\textwidth}@{}}
\toprule
Domain & Short case description ($x$) & Target output ($y^{*}$) & Governing manual ($\mathcal{M}$) and salient cue \\
\midrule
ICD-10-CM clinical coding & Discharge-summary describing metastatic prostate cancer, constipation, aspiration pneumonitis, and chronic comorbidities that must be separated into reportable diagnoses and non-reportable background history. & Exact diagnosis-code set, for example a multi-code answer such as \texttt{K5900}, \texttt{C61}, and related supporting codes. & 2019 ICD-10-CM Guidelines, Alphabetic Index, and Tabular List.\par\smallskip \emph{Manual coding cue.} \texttt{T40.2X5A}: adverse effect of other opioids, initial encounter. \\
\midrule
Federal sentencing guideline calculation & Case involving bank-fraud conspiracy, aggravated identity theft, and marijuana conspiracy, where the grouped counts and the mandatory consecutive count must be handled separately. & Exact total offense level, for example a final grouped offense level of 29. & Title-18 of the United States Code (U.S.C.), United States Sentencing Guidelines (USSG) Appendix A, the offense guideline, and later adjustment chapters.\par\smallskip \emph{Manual Guideline cue.} ``Base offense level: 20'' for 18 U.S.C. \S 1349 fraud conspiracy. \\
\bottomrule
\end{tabular}
\end{table*}

\subsection{ICD-10-CM Clinical Coding}

\paragraph{Description.}
This task evaluates professional diagnosis coding rather than medical question answering. Given a discharge summary, the system must produce the exact diagnosis-code set under ICD-10-CM. Table~\ref{tab:tam-excerpts} gives a short example excerpt, while Appendix~\ref{app:icd-case-study} shows a full worked case with all the steps. The manual is distributed across multiple components: the Alphabetic Index proposes candidate code paths, the Tabular List determines validity and specificity, and the guidelines impose inclusion, exclusion, and additional-coding rules. Rather than resolving the note in a single pass, the coder may need to repeat this lookup cycle across diagnosis candidates. The left panel of Figure~\ref{fig:tam-domain-flows} highlights this repeated diagnosis-level loop. Solving the task therefore requires a multi-stage coding procedure rather than span labeling (like Named Entity Recognition tasks).

\paragraph{Manual complexity.}
The ICD manual is large and interdependent. The 2019 guidelines span 120 pages, the Alphabetic Index spans 1{,}304 pages and contains 7{,}995 main terms, and the Tabular List spans 1{,}942 pages while defining 34{,}502 leaf codes across 21 chapters. These components interact directly: an Index term may trigger a specificity decision in the Tabular List, which may in turn activate a guideline rule that changes the final code set. 

\paragraph{Construction.}
We construct this benchmark from discharge summaries in MIMIC-IV~\cite{johnson2023mimiciv}. The starting slice contains 254{,}377 records of hospital admissions and associated human labeled ICD codes. \rev{R2-3}{As in standard automatic ICD coding benchmarks~\cite{mullenbach-etal-2018-explainable,laat_vu2020,huang-etal-2022-plm}, we treat these coder-assigned labels as the reference target for evaluation.} \rev{R3-1}{Note that MIMIC-IV is public so prior exposure of LLMs to some of its records cannot be ruled out. If such exposure occurred, it would only make the ICD task easier for the model and therefore makes the current low exact-match results conservative.} Additionally, MIMIC-IV cannot be filtered directly to encounters coded under exactly the 2019 ICD-10-CM release. To tackle this, we first restrict to 2017--2019 discharge cases, yielding 65{,}785 candidates centered on that range of years. We then use the official ICD-10-CM yearly addenda and changelogs to identify codes that were added, deleted, or revised across the 2017--2018 and 2018--2019 updates, and exclude any admission containing those revision-sensitive codes. This leaves 53{,}751 admissions whose assigned codes are consistent with a single 2019 manual rather than a mixture of yearly code versions. Finally, because the task input is the discharge summary itself, we require a usable discharge note, producing 38{,}332 final cases of which we sample our evaluation dataset of 1{,}000 cases. 

\paragraph{Data complexity.}
The evaluation sample is a representative subset. It spans 2{,}555 unique output codes, averages 12.85 codes per case, has median 12, p90 23, p99 36, and includes 620 cases with 10 or more codes. These values closely track the full 38{,}332-case cohort, which averages 12.65 codes per case, has median 11, p90 23, p99 34, and includes 23{,}509 cases in the 10+ bucket. Figure~\ref{fig:tam-complexity-bins} (left) shows that the evaluation slice remains dominated by high-cardinality cases. A representative solved ICD example appears in Appendix~\ref{app:icd-case-study}, where the retained answer contains 17 reportable codes.

\subsection{Federal Sentencing Guideline Calculation}

\paragraph{Description.}
Here, the input is a case-fact summary derived from real U.S. federal sentencing materials, and the output is the exact total offense level. The governing sources are the year-aligned Title-18 of the United States Code (U.S.C.), which supplies the federal criminal statutes and related procedure provisions, and the United States Sentencing Guidelines (USSG), which organize sentencing rules into offense-guideline chapters and later chapters for adjustments, grouping, and criminal history. Table~\ref{tab:tam-excerpts} gives a short example excerpt, while Appendix~\ref{app:legal-case-study} shows a full worked case with all the steps. We focus on offense-level calculation rather than final sentencing outcomes because sentencing can also reflect judicial discretion, whereas offense-level computation remains manual-governed and exactly scorable.

A typical path starts from the offense of conviction in Title-18, uses USSG Appendix A to identify the controlling guideline section in the USSG manual, applies the relevant offense characteristics, and then applies later adjustments before computing the total offense level. When a case contains multiple counts or offenses, that process must be repeated and then combined under grouping and cross-reference rules. For example, two counts might first produce offense levels of 30 and 29, then combine to a grouped subtotal of 32 before later reductions.

\paragraph{Manual complexity.}
The legal task has a compact output but a long reasoning horizon. The benchmark spans five annual Title-18 archives and five annual USSG manuals from 2021 through 2025. These manuals contain several guidelines. For instance, the 2024 Title-18 archive contains 1{,}396 section heads, and the 2021 USSG manual spans 608 pages. Difficulty arises from composition: a system can select the wrong statute, miss the Appendix A mapping, apply an unsupported offense characteristic, or fail to recompute the total after grouping and later adjustments. Retrieved snippets may look relevant while still supporting an invalid calculation.

\paragraph{Construction.}
We build this benchmark from federal case materials obtained through CourtListener~\cite{courtlistener2026} in a three-stage curation pipeline. \rev{MR-1}{We first identify candidate dockets from 2021--2025 using document-availability rules rather than sampling dockets arbitrarily. In the current curation snapshot, this yields 833 candidate dockets that contain both at least one eligible sentencing memorandum and at least one non-plea fact-bearing document, while companion plea agreements, factual proffers, and statements of facts are retained when available.} \rev{MR-1}{For each docket, we take the proposed offense level from the selected sentencing memorandum, preferring the government memorandum when available, and use the full document bundle to assemble the supporting case facts.} \rev{MR-1}{We then apply an LLM-based sufficiency audit to discard dockets whose records are too sparse to support reconstructing that offense level; this stage reduces the pool to 250 sufficient dockets.} Appendix~\ref{app:prompt-templates} shows short excerpts from the review prompts used in this pipeline. \rev{MR-1,R1-6}{Finally, we apply human verification to the retained dockets, checking source sufficiency and confirming that the underlying documents support the admitted offense level before admission, yielding the 200 approved cases used in TAM. Accepted cases are therefore not carried forward from a single automated extraction pass alone. This curation strategy intentionally prioritizes source sufficiency and human-verified label quality over raw dataset size.}

\begin{figure}[t]
\centering
\includegraphics[width=\columnwidth]{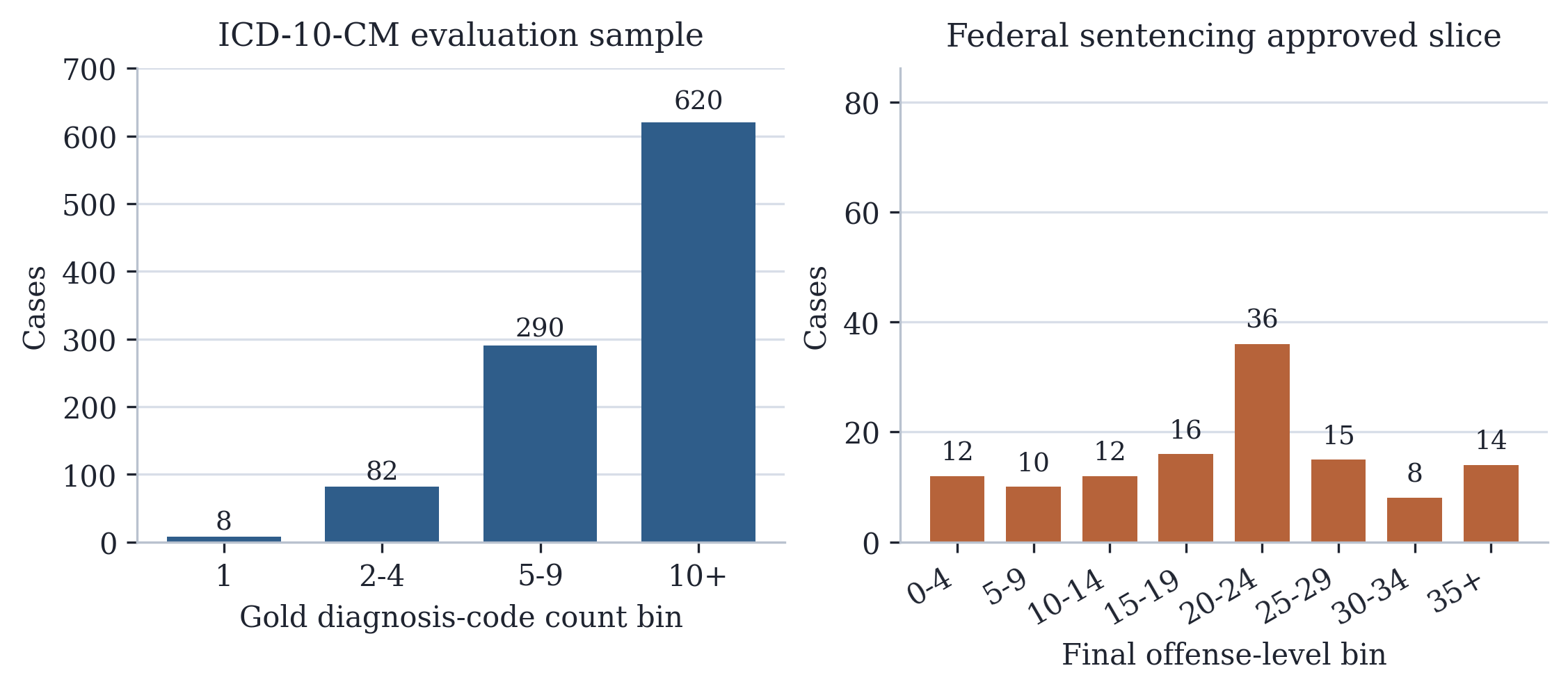}
\caption{Binned output distributions in the ICD evaluation sample (left) and the federal sentencing benchmark (right).}
\Description{Two side-by-side bar charts. The left chart shows ICD-10-CM diagnosis-code count bins for the 1,000-case evaluation sample. The right chart shows offense-level bins for the federal sentencing benchmark.}
\label{fig:tam-complexity-bins}
\end{figure}

\paragraph{Data complexity.}
The legal benchmark is compact in output but dense in evidence. Cases average ~65 extracted facts across an average of 3.07 selected documents. Final offense levels range from 2 to 43, with mean 20.74, median 21, p90 35, and p99 43. Figure~\ref{fig:tam-complexity-bins} (right) shows the distribution across offense-level bins. A representative solved legal example appears in Appendix~\ref{app:legal-case-study}, where the calculation builds separate count-level subtotals, recombines them under grouping, and then applies the final reduction.

\section{Baselines}

We evaluate \agentrev{three} baseline families. All three use the same case inputs and the same manuals. The difference is how the manuals are exposed to the model. Below we explain what each baseline can see and do. Representative prompt excerpts appear in Appendix~\ref{app:prompt-templates}.

\subsection{Retrieval-Augmented Generation Baselines}
We use two retrieval-augmented baselines. The first is a single-pass RAG baseline: it retrieves a small set of manual chunks once, then asks GPT-5 to answer in one shot from the case summary and those snippets. The second is an agentic RAG baseline: it can retrieve, reason, and retrieve again across several rounds before producing a final answer. \rev{R2-4}{In preliminary experiments, more generic RAG setups performed even worse, so the reported variants use light domain-aware chunking and source metadata to make retrieval workable while still remaining prompting-based baselines.}

The ICD RAG variants index the same three sources used throughout the paper: the 2019 Guidelines, the Tabular List, and the Alphabetic Index. We split these manuals in a way that is closer to how an ICD-coder would browse them. Guideline chunks follow section and subsection boundaries, like keeping one chapter or subchapter of a handbook together. Tabular chunks follow code families, so a broad disease category and its more specific sub-diseases stay together unless that branch becomes too large and has to be split. Index chunks follow main terms and nested subterms, which is similar to looking up a disease name and then drilling down into more specific entries underneath it. This process produces 17{,}138 chunks in total: 126 guideline chunks, 2{,}255 Tabular chunks, and 14{,}457 Index chunks.

The legal RAG variants use the same structured chunking idea. We index five annual Title-18 archives and five annual USSG manuals between 2021 and 2025. Title-18 is split by statute section so each criminal statute or procedure provision stays together as one chunk. The USSG is split by guideline section and major subheading so each chunk keeps one offense guideline or one later adjustment, grouping, or criminal-history rule together with the commentary and application notes for that rule. Each legal chunk stores its source year, citation, and semantic path, and retrieval can be filtered to the relevant manual year. This process yields 13{,}091 legal chunks: 6{,}824 from Title-18 and 6{,}267 from the USSG.

The RAG prompts are short and direct. In both domains, the single-pass variant answers after one retrieval round, while the agentic-RAG variant can ask for multiple retrieval rounds when the first set of snippets is not enough. In both domains, the retrieved chunks are formatted with source metadata so the model can see which manual component each snippet came from. Representative prompt excerpts appear in Appendix~\ref{app:prompt-templates}.

\subsection{ReAct (Tool-Using Baseline)}
Our ReAct agent replaces single-pass retrieval with a restricted manual-browsing API. It has to decide which tool to call next, inspect the returned text, and stop when it has enough support for a final answer. This is closer to how a human user works, and it tests whether the model can keep a valid decision path over many steps instead of only summarizing retrieved evidence.

The tool access is domain-specific rather than generic. For ICD-10-CM, the agent receives eight explicit browsing tools over the Alphabetic Index, the Tabular List, and the Official Guidelines. For federal sentencing, the agent receives seven explicit browsing tools over Title-18, Appendix A, and the USSG manual. Appendix~\ref{app:tool-inventory} summarizes the tool roles in each domain.

These tools are deliberately narrow. They expose only the local structure or text the agent asks for: for example, one tool lists candidate Index headings, another opens one exact ICD code, another searches Appendix A by statute citation, and another opens one exact USSG section. The agent still has to decide which manual component to inspect, how to connect the results across tools, and when it has enough support to finalize an answer.

The system prompts for these agents tell the model to act like a careful manual user, stay tied to the case year, prefer inspected manual text over intuition, and return formatted outputs. Representative prompt excerpts and the tool inventory appear in Appendix~\ref{app:prompt-templates} and Appendix~\ref{app:tool-inventory}.

\subsection{Agent Harness Baseline}
\label{sec:agent-harness-baseline}
\agentrev{Our agent-harness baseline leverages LangChain Deep Agents~\cite{langchain_deepagents2026} framework. Unlike ReAct, which keeps the whole procedure inside one running agent loop, the harness breaks the task into a small sequence of specialized workers. Each worker sees only the case materials, manual text, and prior notes needed for its own step, while reusable skills provide compact workflow instructions and domain cues. For ICD-10-CM, these workers handle Index search, Tabular verification, guideline review when needed, and a final coding audit. For federal sentencing, they handle identifying the statute of conviction, mapping it to the relevant sentencing guideline, working through the core offense-level rules, applying later adjustments, and checking the final offense-level calculation. To keep the comparison fair, these workers use the same manual-browsing tools as the ReAct baseline; the main difference is that the harness decomposes the job across staged workers, gives each worker only the local context relevant to its step, and passes concise intermediate reports between them rather than asking one agent loop to carry the whole procedure end to end.}

\section{Experimental Setup}
\label{sec:experimental-setup-placeholder}

\subsection{Evaluation Protocol}

We evaluate all baselines on fixed ICD and sentencing case sets. Each method receives the same case inputs and access to the same underlying manuals, differing only in whether manual access occurs through single-pass retrieval, agentic RAG, tool-based navigation, \agentrev{or an agent harness over the same manual tools}. For ICD-10-CM, the input is a discharge summary and the target is the final diagnosis-code output under the 2019 manual. For federal sentencing, the input is a case-fact summary and the target is the final offense level under the corresponding Title-18 and USSG sources. Each case contributes one final prediction, which is compared against the reference output.

\subsection{Metrics}

Our primary metric in both domains is an exact match. \rev{MR-2,R1-3}{We use exact match as the end-to-end success criterion because each TAM instance has a single manual-governed final output, and a system solves the instance only if it terminates with that exact output. In that sense, exact match is not meant to summarize partial progress; it is the benchmark's strict test of whether the full long-horizon procedure was completed correctly.} Let $y_i$ denote the reference output for case $i$ and $\hat{y}_i$ the model prediction. Exact match over $N$ cases is defined as:
\[
\mathrm{EM} = \frac{1}{N} \sum_{i=1}^{N} \mathbf{1}[\hat{y}_i = y_i].
\]

For ICD-10-CM, this requires the predicted diagnosis-code set to match the reference set exactly, ignoring order; ordering is assessed separately through primary-diagnosis accuracy. For federal sentencing, it requires the predicted offense level to match exactly.

We also report partial-credit metrics. For ICD-10-CM, letting $S_i$ and $\hat{S}_i$ denote the reference and predicted code sets, we report micro-averaged precision and recall:
\[
\mathrm{Precision} = \frac{\sum_{i=1}^{N} |S_i \cap \hat{S}_i|}{\sum_{i=1}^{N} |\hat{S}_i|}, \quad
\mathrm{Recall} = \frac{\sum_{i=1}^{N} |S_i \cap \hat{S}_i|}{\sum_{i=1}^{N} |S_i|},
\]

We also report primary-diagnosis accuracy, which measures whether the first predicted diagnosis code matches the reference primary diagnosis code:
\[
\mathrm{PrimaryDxAcc} = \frac{1}{N} \sum_{i=1}^{N} \mathbf{1}[\hat{y}_{i,1} = y_{i,1}].
\]

For federal sentencing, in addition to exact match, letting $o_i$ and $\hat{o}_i$ denote the reference and predicted offense levels, we report mean absolute error (MAE), which measures the average absolute deviation in offense levels:
\[
\mathrm{MAE} = \frac{1}{N} \sum_{i=1}^{N} |\hat{o}_i - o_i|.
\]
\rev{MR-2,R1-3}{These secondary metrics serve a complementary role: they quantify locally correct partial progress even when the overall trajectory is globally invalid. Interpreting them together with exact match lets us distinguish two questions: whether a model reaches the exact final answer, and how much correct structure it recovers along the way.}

\section{Results}

\subsection{Main results}

\begin{table*}[!t]
\centering
\small
\caption{Main results across baselines. ICD-10-CM exact match stays at 1\% or lower, and legal exact match reaches at most 15.5\%. For ICD-10-CM, EM is exact code-set match, precision and recall measure code-set overlap, and primary-diagnosis accuracy checks whether the first predicted code matches the reference primary diagnosis. For sentencing, EM is exact offense-level match and MAE is mean absolute error in predicted offense levels.}
\label{tab:main-results}
\begin{tabular}{@{}lcccccc@{}}
\toprule
Baseline & ICD EM & ICD Precision & ICD Recall & ICD Prim. Dx Acc. & Legal EM & Legal MAE \\
\midrule
Single-pass RAG & 1\% & 52\% & 33\% & 42.7\% & 7.5\% & 2.87 \\
Agentic RAG & 1\% & 59\% & 37\%  & 45.7\% & 11\% & 2.80 \\
ReAct-style tool use & 1\% & 58.88\% & 26.82\% & 20\% & 15.5\% & 3.04 \\
\agentrev{Agent harness} & \agentrev{0.7\%} & \agentrev{59.9\%} & \agentrev{27.4\%} & \agentrev{50.1\%} & \agentrev{14.5\%} & \agentrev{2.34} \\
\bottomrule
\end{tabular}
\end{table*}

Table~\ref{tab:main-results} reports the main results across baselines. All baseline families remain far from reliable manual execution. \agentrev{On ICD-10-CM, exact match stays at 1\% or below across all baselines, including the agent harness.} On federal sentencing, the best exact match is 15.5\%, achieved by ReAct-style tool use. These results reinforce the central claim of TAM: long-horizon procedural correctness remains difficult even when the model is given retrieval support or a structured tool interface.

On ICD-10-CM, agentic RAG attains the strongest recall among these baselines, reaching 59\% precision, 37\% recall, and 45.7\% primary diagnosis accuracy, but still only 1\% exact match. \agentrev{The agent harness reaches 59.9\% precision and 50.1\% primary diagnosis accuracy, but recall remains 27.4\% and exact match 0.7\%.} \rev{R1-2,R2-3}{Although coder disagreement can lower the absolute ceiling for exact-match evaluation, even the strongest partial ICD results reach only 37\% recall and 50.1\% primary-diagnosis accuracy, suggesting that the gap is unlikely to be explained by label noise alone.} \agentrev{In federal sentencing, the agent harness attains the lowest MAE at 2.34 offense levels, although 35 of 200 cases ended early when retrieved statutory text triggered provider-side content filtering and no response was returned, a behavior we did not observe in the other baselines.} This still corresponds to consequential sentencing differences: under the USSG Sentencing Table, a three-level shift can increase the sentencing range by tens of months, and in high-severity cases by well over five years.

\subsection{Failure Modes}
We analyze 50 saved ReAct trajectories for non-exact cases (25 from ICD-10-CM and 25 from federal sentencing) and label each by its first visible divergence. \rev{R2-2}{We use these labels as a descriptive view of recurring failure patterns rather than a precise estimate of population frequencies.} We group them into three coarse failure modes: global inconsistency, incomplete execution, and missing required elements. 

\paragraph{Global inconsistency.}
\rev{R2-2}{Global inconsistency is the dominant trajectory level failure mode, appearing in 38 of 50 cases overall, including 22 of 25 ICD trajectories and 16 of 25 sentencing trajectories.} In ICD, this usually means that the model selects the wrong anchor term or code family early and then continues coherently within the wrong diagnostic branch. In sentencing, it appears as either a wrong guideline anchor or an aggregation step performed without a global state check. \rev{R1-4}{In many cases, the system appears to enter a plausible path through the manual and then lose consistency as later decisions accumulate. Because retrieval and reasoning are tightly coupled in TAM, we do not try to separate them in this analysis.}

\paragraph{Incomplete execution.}
\rev{R2-2}{Incomplete execution appears in 7 of 50 trajectories overall, including 3 of 25 ICD trajectories and 4 of 25 sentencing trajectories.} In ICD, this appears when the model stops before a full secondary-diagnosis sweep; in sentencing, it appears as an incomplete trajectory that never carries the calculation through to a complete offense-level determination. In both domains, the model begins a plausible procedure but terminates before it verifies global completeness.

\paragraph{Missing required elements.}
\rev{R2-2}{Missing required elements appear in 5 of 50 trajectories overall and only in sentencing, where they appear in 5 of 25 cases.} These failures arise when the model misses a mandatory reduction or slips on a late-step required constraint after otherwise entering a plausible guideline path. We do not observe this pattern as the first visible divergence in the current ICD trajectories sample, suggesting that ICD trajectories more often fail earlier by entering the wrong branch altogether.

\section{Future Research}
TAM points to several directions for improving long-horizon procedural reasoning: stronger state tracking, explicit verification before finalization, and mechanisms for revision or backtracking after early mistakes. \rev{MR-3,R1-5,R2-1,R3-3}{A natural next step is broader evaluation with stronger domain-specialized systems, fine-tuned models, and future agentic methods as the community adopts TAM.} Broadly, these directions suggest a need for methods that treat reasoning as a structured, stateful process rather than a sequence of loosely connected steps.

\section{Limitations}

TAM focuses on two domains with well-defined manuals and exact outputs, which enables precise evaluation but does not cover tasks with looser procedures or heavier reliance on human judgment. We also evaluate representative prompting-based baselines rather than an exhaustive set of specialized or trained systems.

\section{Conclusion}

TAM exposes a gap between short-horizon reasoning benchmarks and real-world tasks that require following large application manuals step by step. Across ICD-10-CM coding and federal sentencing, current retrieval-based and tool-using approaches often recover plausible local decisions but fail to execute the full procedure correctly.
These findings suggest that access to relevant information is not enough: the harder problem is maintaining consistency, satisfying constraints, and completing the full manual-governed decision process end to end. We hope TAM helps measure that capability and motivates methods built for long-horizon procedural execution.

\appendix

\section{Appendix Overview}
This appendix provides supplementary prompt excerpts, tool inventories, and condensed case traces referenced in the main text. Appendix~\ref{app:prompt-templates} presents clipped excerpts from the prompts used for the single-pass RAG, agentic-RAG, and ReAct, together with a short harness overview and the ReAct tool inventory. Appendix~\ref{app:icd-case-study} and Appendix~\ref{app:legal-case-study} provide condensed worked traces for one ICD-10-CM case and one federal sentencing case, illustrating the end-to-end manual-following process.

\section{Prompt Templates and Agent Instructions}
\label{app:prompt-templates}

We quote clipped stretches of the real prompts here that illustrate the task framing, workflow instructions, grounding rules, and expected outputs that each method actually sees. Omitted placeholders, repeated schema fields, and boilerplate are marked as ``(...)''.

\subsection{RAG and Agentic-RAG Prompts}

\paragraph{ICD single-pass RAG.}
\begin{tampromptexcerpt}
``Task: read the case summary, use the retrieved ICD manual snippets as supporting context, and predict all diagnosis ICD-10-CM codes explicitly supported by the summary, not just the primary diagnosis. (...) Rules: use only the case summary and retrieved snippets, do not invent conditions that are not stated or strongly implied, prefer conservative omission over guessing, and leave out codes whose support is unclear. (...) Output: \texttt{predicted\_icd\_codes}, confidence, rationale, supporting evidence.''
\end{tampromptexcerpt}

\paragraph{Legal single-pass RAG.}
\begin{tampromptexcerpt}
``Task: read the case summary, use the retrieved legal context, and predict the most defensible total offense level; if the inspected context clearly supports them, also predict the criminal history category and guideline range. (...) Rules: use only the case summary and retrieved legal context, prefer omission over speculation, and return null for unsupported fields instead of filling them from general legal knowledge. (...) Output: offense level, criminal history category, guideline range, confidence, rationale, supporting evidence.''
\end{tampromptexcerpt}

\paragraph{ICD agentic RAG.}
\begin{tampromptexcerpt}
``Task: use only the ICD retrieval tools; one full-case search is allowed for a broad first pass over the entire summary, followed by focused searches for narrower diagnoses, code families, or rule checks. Predict all explicitly supported diagnosis codes, not just the primary diagnosis. (...) Rules: use only the case summary and retrieved ICD snippets, do not invent unstated conditions, prefer conservative omission over guessing, and leave out unclear codes. (...) Output: predicted ICD code list with confidence, rationale, and supporting evidence.''
\end{tampromptexcerpt}

\paragraph{Legal agentic RAG.}
\begin{tampromptexcerpt}
``Task: use only the legal retrieval tools and the correct yearly edition; one full-case search is allowed for a broad first pass over the case summary, followed by focused searches for statutes, guideline sections, or adjustments. Predict the most defensible total offense level and, when clearly supported, the criminal history category and guideline range. (...) Rules: use the USSG for offense-level calculation and Title-18 only when statutory text is needed, stop once support is sufficient, and return null for unsupported fields rather than extrapolating past the evidence. (...) Output: total offense level, criminal history category, guideline range, confidence, rationale, supporting evidence.''
\end{tampromptexcerpt}

\subsection{ReAct/Manual-Navigation Prompts}

\paragraph{ICD ReAct prompt excerpt.}
\begin{tampromptexcerpt}
``Required workflow: start from diagnoses, symptoms, and chronic conditions that were clearly active for the encounter; use the Alphabetic Index to find lead terms and candidate code families; confirm every retained code in the Tabular List; and consult the Official Guidelines whenever sequencing, exclusion, combination, specificity, or additional-code rules may change the final answer. (...) Keep only the final compliant ICD-10-CM diagnosis code set. Remove unsupported codes, redundant symptom codes, and code combinations blocked by guideline or Tabular logic. Before answering, do one final completeness sweep for active secondary diagnoses and separately codeable findings that clearly affected the encounter. (...) Coding rules: never finalize a code without inspecting the exact Tabular entry; prefer the most specific supported code or supported combination code; do not separately code symptoms integral to a confirmed diagnosis unless the manual text supports doing so.''
\end{tampromptexcerpt}

\paragraph{Legal ReAct prompt excerpt.}
\begin{tampromptexcerpt}
``Core stance: work like a careful legal manual user, stay grounded in the case year, and inspect controlling text before making any offense-level claim. (...) Required workflow: identify the statute of conviction, map it through Appendix A to the Chapter Two guideline, determine the Chapter Two offense level from the inspected guideline text, and then apply Chapter Three adjustments supported by the case facts and inspected manual text. Do not skip the Appendix A mapping step. If Appendix A returns multiple candidate guideline sections, inspect the candidate sections and choose only the section supported by the statute and case facts. (...) Decision rules: do not choose a guideline from memory, do not infer a final total offense level solely because an adjustment is common or likely, and return null when a material step lacks manual and factual support.''
\end{tampromptexcerpt}

\subsection{Agent-Harness Overview}

\agentrev{As described in Section~\ref{sec:agent-harness-baseline}, we implement the agent-harness baseline with LangChain Deep Agents~\cite{langchain_deepagents2026}. Unlike the ReAct baseline, which keeps the full trajectory inside one growing conversation, the harness separates the work into staged workers and reusable skills. Each worker sees only the case materials, manual text, and prior notes needed for its own step, writes a short structured report, and hands that report to the next worker. Full worker prompts, skill files, and runtime configuration are provided in the public repository.}

\subsection{Construction-Time Review Prompts}

These prompts are used during federal benchmark curation to verify that the case record is sufficient to support the proposed offense level and extract the case facts.

\paragraph{Offense-level verification prompt.}
\begin{tampromptexcerpt}
``Primary objective: convert one docket bundle into a structured offense-level review artifact (...) Grounding rules: use only the docket bundle and current tool outputs, prefer direct document evidence over legal intuition, surface incompleteness explicitly, and return null when the total offense level is not supportable. (...) Workflow: identify the strongest sentencing documents, derive the offense-level calculation from sentencing information first, then substantiate each step with docket facts while keeping the claim, sentencing evidence, guideline support, and justification separately legible. (...) Returned artifact: docket id, support status, selected documents, case facts, offense-level steps, final total offense level.''
\end{tampromptexcerpt}

\paragraph{Case-facts extraction prompt.}
\begin{tampromptexcerpt}
``Return only valid JSON with a \texttt{case\_facts} list. (...) Extraction rules: extract all grounded case facts in the docket bundle, not just those used in the offense-level computation; prefer high recall over brevity; keep facts atomic and concrete; and include procedural facts, charge facts, plea facts, admitted conduct, loss figures, and other grounded details needed for downstream reconstruction. (...) Exclude guideline rules, offense-level arithmetic, and personally identifying information, and rewrite mixed fact-plus-sentencing sentences so that only the underlying factual content remains.''
\end{tampromptexcerpt}

\subsection{ReAct Tool Summary}
\label{app:tool-inventory}

The ReAct baseline receives explicit domain-specific manual-browsing tools in each domain (full implementations are in the public repository).

For \textbf{ICD-10-CM}, there are eight tools in total: two for the Alphabetic Index, four for the Tabular List, and two for the Official Guidelines. Together they let the agent list top-level Index headings, open one Index hierarchy, move through Tabular chapters and blocks, inspect one exact ICD code, list the guideline table of contents, and open one exact guideline section.

For \textbf{federal sentencing}, there are seven tools in total: three for Title-18 browsing, one for Appendix A statute-to-guideline mappings, and three for the USSG. Together they let the agent list and open Title-18 chapters, inspect one exact statute section, list the USSG table of contents, open one USSG subheading, search Appendix A mappings, and open one exact USSG guideline section with commentary.

\section{Additional Examples and Case Studies}
We include two condensed traces drawn from real benchmark cases. The examples omit repetitive lookup steps while preserving the key procedural decisions and governing manual path.

\subsection{ICD-10-CM Case Study}
\label{app:icd-case-study}

\paragraph{Case setup and summary.}
This ICD case is an admission for progressive abdominal pain, vomiting, and four days without a bowel movement in a patient with metastatic prostate cancer and multiple chronic neurologic comorbidities. The final reference code set contains 17 diagnosis codes, illustrating repeated diagnosis-by-diagnosis lookup loop. The discharge summary  describes abdominal pain, vomiting, constipation, and focal sigmoid narrowing in a patient with newly diagnosed metastatic prostate cancer, while the discharge diagnosis explicitly identifies opioid-induced constipation with concurrent prostate cancer and hypertension. The retained code set then expands across cancer, bone metastasis, neurologic complications, aspiration pneumonitis, chronic comorbidities, medication adverse effect, pain, weakness, and status-history coding. 

\paragraph{Representative manual cue.}
\begin{tampromptexcerpt}
ICD-10-CM supplies explicit navigation cues rather than leaving the coder to invent the path. For example, the manual can require a separate adverse-effect code such as \texttt{T40.2X5A} for opioid-related constipation instead of letting that fact remain implicit in the narrative.
\end{tampromptexcerpt}
These short instructions show why ICD coding cannot stop at identifying diagnoses in the note: the manual can require opening another code path before the answer is complete. The steps below sketch that coding path:

\begin{enumerate}
  \item Start from the documented problems in the note, but use the Alphabetic Index to choose the next lead term.
  \item Follow the constipation path first: the Index proposes the constipation code family, and the Tabular entry determines the retained constipation code and whether it is reportable for this encounter.
  \item When the note ties constipation to opioids, follow the manual's adverse-effect instruction to open a separate drug-related code path instead of folding that fact into the constipation code.
  \item Resolve the malignancy path next: use the Index and Tabular List to code the primary prostate cancer and then the secondary bone metastasis, with Tabular notes deciding whether the neoplasm codes trigger sequencing or dependency constraints.
  \item Process the remaining active diagnoses one manual path at a time, including the neurologic conditions, aspiration pneumonitis, and chronic comorbidities, because each retained diagnosis requires its own Index entry and Tabular verification.
  \item Use the guideline and Tabular instructions to decide whether status/history items such as long-term insulin use and prior intracerebral hemorrhage belong in the final answer, rather than treating them as ordinary active diagnoses.
  \item Only after the manual has resolved each candidate path do you assemble, de-duplicate, and order the retained codes, checking that the final output contains exactly 17 reportable codes.
  \item Report the 17 codes: K5900, J690, G9341, C7951, F0391, F05, R4701, C61, G40909, E119, T402X5A, G893, R531, I10, E785, Z794, Z8673

\end{enumerate}

\subsection{Federal Sentencing Case Study}
\label{app:legal-case-study}

\paragraph{Case setup and summary.}
This legal case involves three offenses: conspiracy to commit bank fraud, aggravated identity theft, and conspiracy to manufacture or distribute marijuana. The factual basis describes a credit-card bust-out scheme using stolen and synthetic identities and staged payments or chargebacks, with fraud proceeds then used to operate a marijuana grow and retail business. The case involves two different substantive count groups with different guideline anchors, plus a separate aggravated identity-theft count that stays outside the grouped offense-level arithmetic.

\paragraph{Representative manual cue.}
\begin{tampromptexcerpt}
USSG \S 3D1.4 grouping table: total of 2 Units $\rightarrow$ add 2 levels.
\end{tampromptexcerpt}
This kind of short rule shows why the legal task cannot stop at separate count calculations: the final answer depends on recombining them under the grouping rules. The steps below sketch that manual path:

\begin{enumerate}
  \item Work count by count: Count 1 is bank-fraud conspiracy, Count 2 is aggravated identity theft, and Count 3 is marijuana conspiracy.
  \item Fix the manual year so the statute text, Appendix A mapping, and guideline sections all come from the same edition.
  \item For Count 1, use USSG \S\S 2X1.1 and 2B1.1. Start at base offense level 7, then add +16 for loss, +2 for victim count, +2 for sophisticated means, and +3 for role, for a subtotal of 30.
  \item For Count 3, switch to USSG \S 2D1.1. Start at base offense level 24, then add +2 for maintaining a drug premises and +3 for role, for a subtotal of 29.
  \item Keep Count 2 separate: aggravated identity theft under 18 U.S.C. \S 1028A does not enter the grouped offense-level calculation.
  \item Under USSG \S 3D1.4, the higher 30-level group counts as 1 Unit, and the 29-level group adds 1 more Unit because it is only one level lower.
  \item The grouping table maps 2 Units to +2 levels, so the grouped subtotal rises from 30 to 32.
  \item Apply the acceptance-of-responsibility reduction of -3 under USSG \S 3E1.1.
  \item Report the final grouped offense level as 29. Count 2 remains outside that arithmetic.
\end{enumerate}

\section*{GenAI Usage Disclosure}
Generative AI tools were used as evaluated systems in this study and for limited drafting and editing assistance during manuscript preparation. 

\bibliographystyle{ACM-Reference-Format}
\balance
\bibliography{tam-references}

\end{document}